\documentclass{article}
\usepackage{microtype}
\usepackage{graphicx}
\usepackage{booktabs} 
\usepackage{amsfonts}
\usepackage{amsmath}
\usepackage{listings}
\usepackage{xcolor}
\usepackage{float}
\usepackage{tabularx} 
\usepackage{caption}
\usepackage{subcaption}
\usepackage{adjustbox}
\usepackage{enumitem}
\usepackage{tikz}
\usepackage{pict2e,circledsteps,pgfkeys}

\usepackage{hyperref}
\usepackage{array}
\usepackage{colortbl}
\usepackage{xurl}
\usepackage{color,soul}
\usepackage{xspace}

\definecolor{pblue}{RGB}{78,121,167}
\definecolor{pred}{RGB}{225,87,89}
\definecolor{pgreen}{RGB}{89,161,79}
\definecolor{darkgreen}{rgb}{0,0.5,0}

\newcommand*\bcircled[1]{\tikz[baseline=(char.base)]{
            \node[shape=circle,fill,inner sep=0.5pt] (char) {\textcolor{white}{\textit{#1}}};}}

\newcommand*\wcircled[1]{\tikz[baseline=(char.base)]{
            \node[shape=circle,draw=black,inner sep=0.5pt] (char) {\textcolor{black}{\textit{#1}}};}}

\lstdefinestyle{xmlstyle}{
  basicstyle=\ttfamily\footnotesize,
  morestring=[s]{"}{"},
  morecomment=[s]{?}{?},
  morecomment=[s]{!--}{--},
  commentstyle=\color{darkgreen},
  moredelim=[s][\color{black}]{>}{<},
  moredelim=[s][\color{pred}]{\ }{=},
  stringstyle=\color{pgreen},
  identifierstyle=\color{pblue},
  tabsize=2,
  breaklines=true,
  frame=single,
  xleftmargin=3.4pt,
  xrightmargin=3.4pt
}

\usepackage[accepted]{mlsys2024}

\mlsystitlerunning{\tech: Modular Attention Reuse for Low-Latency Inference}

\begin{document}

\definecolor{lightgray}{gray}{0.9}
\definecolor{lightblue}{rgb}{0.9,0.9,1}
\definecolor{blue_bg}{rgb}{0.85,0.85,1}
\definecolor{lightyellow}{rgb}{1,1,0.8}
\definecolor{lightpurple}{rgb}{1,0.85,1}
\definecolor{red}{rgb}{1,0,0}
\definecolor{darkgreen}{rgb}{0.4,0.7,0.3}
\definecolor{darkblue}{rgb}{0.2,0.7,0.9}

\newcommand\note[1]{\hl{-- #1 --}} 
\newcommand{\hlc}[2][yellow]{ {\sethlcolor{#1} \hl{#2}} }
\newcommand\lin[1]{\hlc[yellow]{LZ: -- #1 --}} 
\newcommand\nikhil[1]{\hlc[darkblue]{NS: -- #1 --}} 
\newcommand\lina[1]{\hlc[gray]{LZ: -- #1 --}} 
\newcommand\slee[1]{\hlc[blue_bg]{SL: -- #1 --}} 
\newcommand\sleea[1]{\hlc[gray]{SL: -- #1 --}} 
\newcommand\gj[1]{\hlc[orange]{GJ: -- #1 --}} 
\newcommand{\code}[1]{{\texttt{\small #1}}}
\newcommand{\program}[1]{\textsf{#1}}

\newcommand\notes[1]{{\color{violet} #1}}

\newcommand{\pt}[1]{{\textbf{\textsf{\scriptsize{#1}}}}}


\newcommand{\system}{PromptCache\xspace}
\newcommand{\tech}{Prompt Cache\xspace}
\newcommand{\Sch}{Schema\xspace}
\newcommand{\sch}{schema\xspace}
\newcommand{\schs}{schemas\xspace}

\newcommand{\modu}{prompt module\xspace}
\newcommand{\modus}{prompt modules\xspace}
\newcommand{\kvcache}{KV Cache\xspace}
\newcommand{\lang}{PML\xspace}

\newcommand{\ie}{\textit{i.e.}\xspace}
\newcommand{\eg}{\textit{e.g.}\xspace}


\newcommand{\papertitle}{\tech: Modular Attention Reuse\\\ for Low-Latency Inference}

\twocolumn[
\mlsystitle{\papertitle}



\mlsyssetsymbol{equal}{*}

\begin{mlsysauthorlist}
    \mlsysauthor{In Gim}{yale}
    \mlsysauthor{Guojun Chen}{yale}
    \mlsysauthor{Seung-seob Lee}{yale}
    \mlsysauthor{Nikhil Sarda}{google}
    \mlsysauthor{Anurag Khandelwal}{yale}
    \mlsysauthor{Lin Zhong}{yale}
\end{mlsysauthorlist}
    
\mlsysaffiliation{yale}{Department of Computer Science, Yale University, USA. \{in.gim, guojun.chen, seung-seob.lee, anurag.khandelwal, lin.zhong\}@yale.edu\ }
\mlsysaffiliation{google}{Google, Mountain View, California, USA. nikhilsarda@google.com}

\mlsyscorrespondingauthor{Lin Zhong}{lin.zhong@yale.edu}

\mlsyskeywords{Machine Learning, MLSys}

\vskip 0.3in

\begin{abstract}
    We present \textit{\tech}, an approach for accelerating inference for large language models (LLM) by reusing attention states across different LLM prompts. Many input prompts have overlapping text segments, such as system messages, prompt templates, and documents provided for context.
Our key insight is that by precomputing and storing the attention states of these frequently occurring text segments on the inference server, we can efficiently reuse them when these segments appear in user prompts. \tech employs a \textit{\sch} to explicitly define such reusable text segments, called \modus.  The \sch ensures positional accuracy during attention state reuse and provides users with an interface to access cached states in their prompt.
Using a prototype implementation, we evaluate \tech across several LLMs. We show that \tech significantly reduce latency in time-to-first-token, especially for longer prompts such as document-based question answering and recommendations. The improvements range from \textbf{8$\times$} for GPU-based inference to \textbf{60$\times$} for CPU-based inference, all while maintaining output accuracy and without the need for model parameter modifications.

\end{abstract}
]



\printAffiliationsAndNotice{}  
\section{Introduction}
\label{introduction}

A substantial fraction of large language model (LLM) prompts are reused frequently. For example, prompts usually commence with identical ``system messages'' that provide initial guidelines for its functionality. 
Documents can also overlap in multiple prompts. In a wide range of long-context LLM applications, such as legal analysis~\cite{cui2023chatlaw, nay2023large}, healthcare applications~\cite{DBLP:journals/jbi/SteinbergJFCPS21,DBLP:journals/npjdm/Rasmy0XTZ21}, and education~\cite{DBLP:journals/corr/abs-2106-07340}, the prompt includes one or several documents from a pool. 
Additionally, prompts are often formatted with reusable templates~\cite{DBLP:journals/corr/abs-2302-11382} as a result of prompt engineering. Such examples are common in LLM for robotics and tool learning~\cite{huang2022language, driess2023palme, DBLP:journals/corr/abs-2304-08354}. This further results in a high degree of overlap between prompts using the same template.

We introduce a novel technique termed \emph{\tech} to reduce the computational overhead in generative LLM inference. \tech is motivated by the observation that input prompts to LLM often has reusable structures. The key idea is to precompute attention states of the frequently revisited prompt segments in memory, and reuse them when these segments appear in the prompt to reduce latency.

\begin{figure*}
    \centering
    \subcaptionbox{Autoregressive token generation\label{fig:overview-sub1}}[0.29\textwidth]{\includegraphics[width=0.25\textwidth]{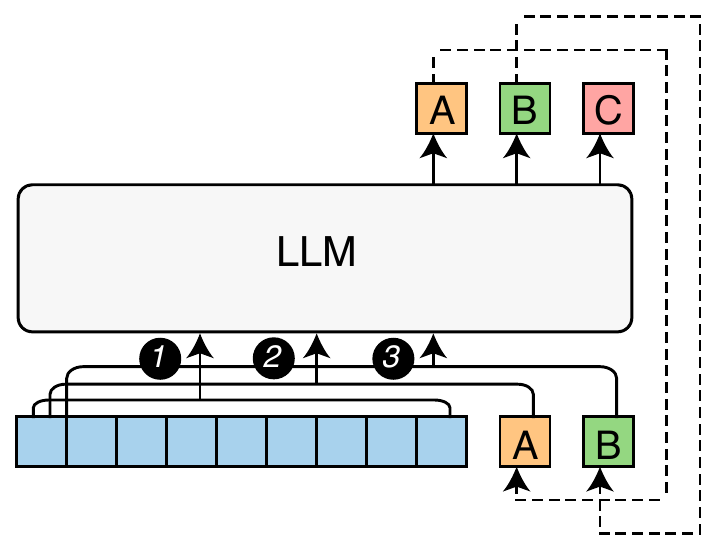}}
    \subcaptionbox{Generation with \kvcache\label{fig:overview-sub2}}[0.29\textwidth]{\includegraphics[width=0.25\textwidth]{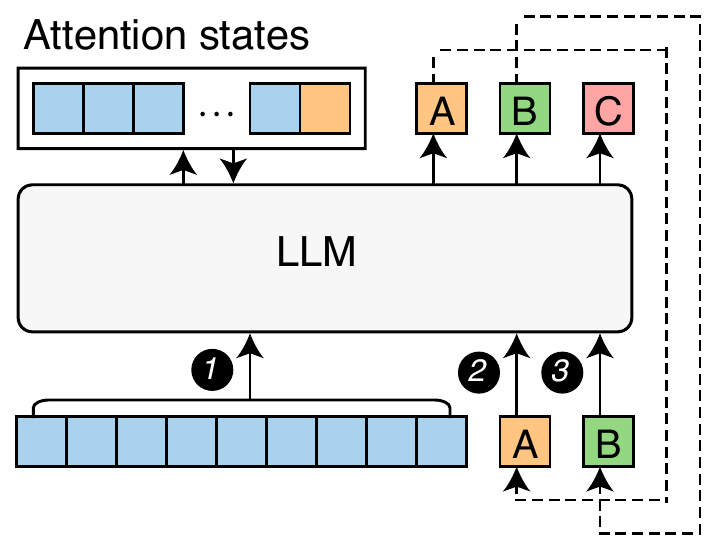}}
    \subcaptionbox{Generation with \tech\label{fig:overview-sub3}}[0.41\textwidth]{\includegraphics[width=0.35\textwidth]{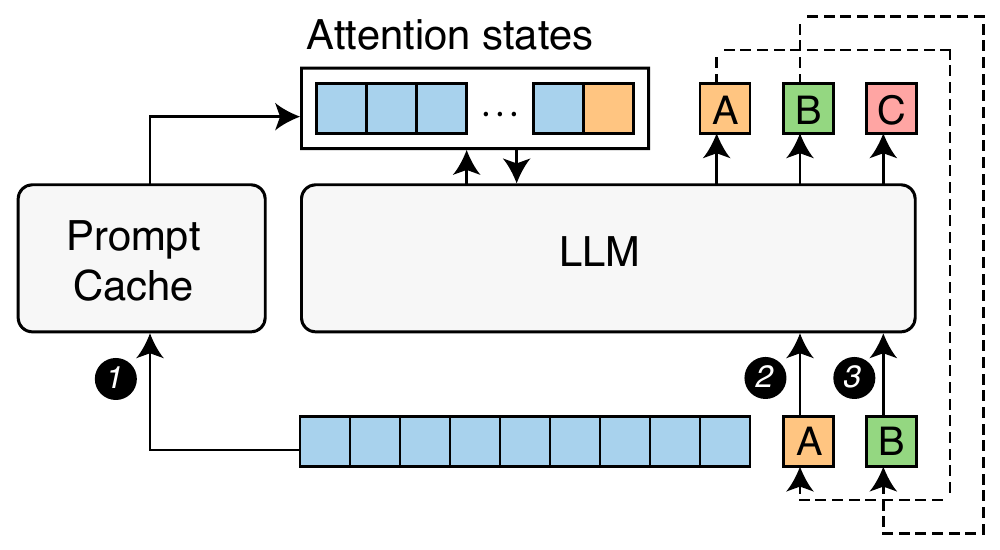}}
    \caption{
    Comparison of LLM token generation methods, each showing three steps (\bcircled{1}  to \bcircled{3}). Each box indicates a token. Blue boxes represent the prompt.  (a) An LLM takes in a prompt (blue tokens) and predicts the next token ({\tiny \boxed{A}}) (\bcircled{1}). It then appends the generated token ({\tiny \boxed{A}}) to the prompt to predict the next token ({\tiny \boxed{B}}) (\bcircled{2}). This process, called autoregressive, continues until a stop condition is met.  (b) \kvcache computes time attention states for the prompt only once (\bcircled{1}) and reuses them in the following steps; (c) \tech reuses the KV state across services to bypass prompt attention computation. 
    \tech populates its cache when a \sch is loaded and reuses the cached states for prompts that are derived from the \sch (\bcircled{1}). \autoref{fig:schema-prompt} further elaborates Step \bcircled{1}. 
    }  
     \label{fig:overview}
\end{figure*}

Reusing attention states is a popular strategy for accelerating the service of a single prompt~\cite{pope2022efficiently}. The existing approach, often referred to as \emph{Key-Value (KV) Cache}, reuses the key-value attention states of input tokens during the autoregressive token generation.
This eliminates the need to compute full attention for every token generation (\S~\ref{ssec:kvcache}). By caching the key-value attention computed for the previously generated token, each token generation requires the computation of key-value attention states only once. 

Building on top of \kvcache, \tech extends attention state reuse from a single prompt to multiple prompts by making attention state reuse \emph{modular}.
In our approach, frequently reused text segments are individually precomputed and stored in memory. When such ``cached'' segments appear in the input prompt, the system uses the precomputed key-value attention states from memory instead of recomputing them. As a result, attention computations are only required for uncached text segments. \autoref{fig:overview} illustrates the difference between full autoregressive generation, \kvcache, and \tech. We note that the performance advantage becomes more pronounced as the size of cached segments grows since the computation overhead of attention states scales \emph{quadratically} with input sequence size~\cite{keles2022computational,tay2023efficient} while the space and compute complexity of \tech scales \emph{linearly} with the size.

Two challenges arise when reusing attention states across prompts. First, attention states are position-dependent due to the positional encoding in Transformers. Thus, the attention states of a text segment can only be reused if the segment appears at the same position. Second, 
the system must be able to efficiently recognize a text segment whose attention states may have been cached in order to reuse.

To tackle these two problems, \tech combines two ideas.
The first is to make the structure of a prompt explicit with a \emph{Prompt Markup Language} (\lang).
\lang makes reusable text segments explicit as modules, \ie, \emph{\modu}. 
It not only solves the second problem above but opens the door for solving the first, since each \modu can be assigned with unique position IDs.
Our second idea is our empirical finding that LLMs can operate on attention states with discontinuous position IDs. This means that we can extract different segment of attention states and concatenate them to formulate subset of meanings. We leverage this to enable users to select \modus based on their needs, or even update some \modus during the runtime.

We explain how \tech works in \S\ref{sec:prompt-caching}. In summary, an LLM user writes their prompts in \lang, with the intention that they may reuse the attention states based on \modus. 
Importantly, they must derive a prompt from a \emph{\sch}, which is also written in \lang. \autoref{fig:schema-prompt} shows a example prompt based on an example \sch.
When \tech receives a prompt, it first processes its \sch and computes the attention states for its \modus. It reuses these states for the \modus in the prompt and other prompts derived from the same \sch.
In \S\ref{sec:implementation}, we report a prototype implementation of \tech on top of the HuggingFace transformers library \cite{wolf2020huggingfaces}. While \tech can work with any Transformer architecture compatible with \kvcache, we experiment with three popular Transformer architectures powering the following open-sourced LLMs: Llama2 \cite{touvron2023llama}, Falcon \cite{falcon}, and MPT~\cite{MosaicML2023Introducing}. We consider two types of memory for storing prompt modules: CPU and GPU memory. While CPU memory can scale to terabyte levels, it brings the overhead of host-to-device memory copying. In contrast, GPU memory does not require coping but has limited capacity.

Using the prototype, we conduct an extensive benchmark evaluation to examine the performance and quantify the accuracy of \tech across various long-context datasets (\S\ref{sec:eval}). We employ the LongBench suite~\cite{bai2023longbench}, which includes recommendation and question-answering (QA) tasks based on multiple documents. In our evaluation, \tech reduces time-to-first-token (TTFT) latency from $1.5\times$ to $10\times$ for GPU inference with prompt modules on GPU memory and from $20\times$ to $70\times$ for CPU inference, all without any significant accuracy loss. Additionally, we analyze the memory overhead of the precomputed attention states for each model and discuss directions for optimizing the memory footprint of \tech. 
We subsequently showcase several generative tasks, including personalization, code generation, and parameterized prompts, to demonstrate the expressiveness of the prompt schema and performance improvement with negligible quality degradation.

In our present study, we mainly focus on techniques for modular attention reuse. However, we foresee \tech being utilized as a foundational component for future LLM serving systems. Such systems could incorporate enhanced prompt module management and GPU cache replacement strategies, optimizing the advantages of both host DRAM and GPU HBM. Our source code and data used for evaluation are available at \href{https://github.com/yale-sys/prompt-cache}{github.com/yale-sys/prompt-cache}.

\section{Background and Related Work}

\tech builds on the ideas of the \kvcache, \ie, key-value attention state reuse during autoregressive decoding in LLMs. This section reviews autoregressive token generation in LLMs, explains how the incorporation of \kvcache can speed up the token generation process, identifies its approximations, and surveys recent work that leverages the \kvcache for acceleration. We also briefly discuss other existing techniques for accelerating LLM inference.

\subsection{Autoregressive Token Generation}

An LLM generates output tokens autoregressively~\cite{radford2018improving}. It starts with an initial input, often called a prompt, and generates the next token based on the prompt. The model then appends the token to the prompt and uses it to generate the next token. 
The generation process continues until a stopping condition is met. This could be after a predetermined number of tokens, upon generating a special end-of-sequence token, or when the generated sequence reaches a satisfactory level of coherence or completeness.
Importantly, in each step, the model takes the entire prompt and tokens generated so far as the input, and repeat.

\subsection{Key-Value Cache}
\label{ssec:kvcache}

Autoregressive token generation described above incurs substantial computation due to the self-attention mechanism being applied over the entirety of input during each step. To ameliorate this, the Key-Value (KV) Cache mechanism~\cite{pope2022efficiently} is frequently used. This technique computes the key and value embeddings for each token only once throughout the autoregressive token generation. To elaborate, denote a user prompt as a sequence of $n$ tokens: $s_1,\dots, s_n$, and the subsequently generated $k$ tokens as $s_{n+1},\dots, s_{n+k}$. In naive autoregressive token generation, the attention states $\{(k_1, v_1), \dots, (k_{n+k}, v_{n+k})\}$ are fully recalculated at every step.
In contrast, \kvcache initially computes attention states for the input, represented by $S_0 = \{(k_i, v_i) | i\leq n\}$, and caches them in memory. This step is often referred to as the \textit{prefill} phase.
For every subsequent step $j\leq k$, the model reuses the cached values $S_j = \{(k_i, v_i) | i< n + j\}$ to compute the attention state $(k_{n+j}, v_{n+j})$ of the new token $s_{n+j}$. This approach significantly reduces the computation required for self-attention. Specifically, the computation in each step, measured in FLOPs for matrix operations, is reduced by a factor of $1/n$. The number of operations decreases from approximately $6nd^2+4n^2d$ to $6d^2+4nd$, where $d$ is a hidden dimension size.  After each step, the newly computed $(k_{n+j}, v_{n+j})$ attention states are appended to the cache for subsequent use. In causal language models, which account for most LLMs, the use of \kvcache does not affect the model's accuracy, since the attention at position $i$ is computed based solely on the tokens at positions located before $i$-th token.

The \kvcache has catalyzed further exploration into LLM acceleration. Ensuing studies have either centered on refining memory management for \kvcache, as demonstrated in \textit{paged attention}~\cite{kwon2023efficient}, on pruning superfluous \kvcache data~\cite{zhang2023h2o}, or compressing it~\cite{liu2023cachegen}. There are some preliminary works that explore \kvcache reuse across different requests as well. \cite{feng2023attmemo} reuse memorized attention states based on an embedding similarity metric. Paged attention also demonstrates simple prefix sharing, where different prompts with an identical prefix share \kvcache. However, existing approaches are specific to certain scenarios, while we investigate attention reuse for \emph{general} LLM prompts.

\subsection{Other Methods for Low-Latency LLM Inference}

\tech introduces an orthogonal optimization strategy that augments existing systems dedicated to efficient LLM inference. This includes systems that utilize multiple GPUs for inference~\cite{aminabadi2022deepspeed} and those with high-performance GPU kernels for softmax attention score computation~\cite{dao2022flashattention}. Although our current focus is on achieving low-latency inference in LLMs, \tech can also benefit systems aiming for high throughput~\cite{sheng2023flexgen} as well via reduced computation.
\section{Design of \tech}
\label{sec:prompt-caching}
The effectiveness of the \kvcache leads us to the next question: \textit{Can attention states be reused across multiple inference requests?} We observe that different prompts often have overlapping text segments. For example, identical ``system messages", or metaprompts are frequently inserted at the beginning of a prompt to elicit desired responses from an LLM.
For another example, in many legal and medical applications of LLMs~\cite{cui2023chatlaw, DBLP:journals/jbi/SteinbergJFCPS21,DBLP:journals/npjdm/Rasmy0XTZ21}, the same set of documents is often provided as context to different prompts. 
Finally, reusable prompt formats, \ie, \textit{prompt templates}, are commonly used by LLM applications in robotics and tool learning~\cite{driess2023palme, DBLP:journals/corr/abs-2304-08354}, since most tasks are variations of a few common task. In this section, we describe our approach called \emph{\tech}, which answers the above question affirmatively. \tech improves computational efficiency through \textit{inter-request} attention state reuse by leveraging the shared segments in a structured manner.

\begin{figure*}[t]
\centering

\includegraphics[width=0.9\textwidth]{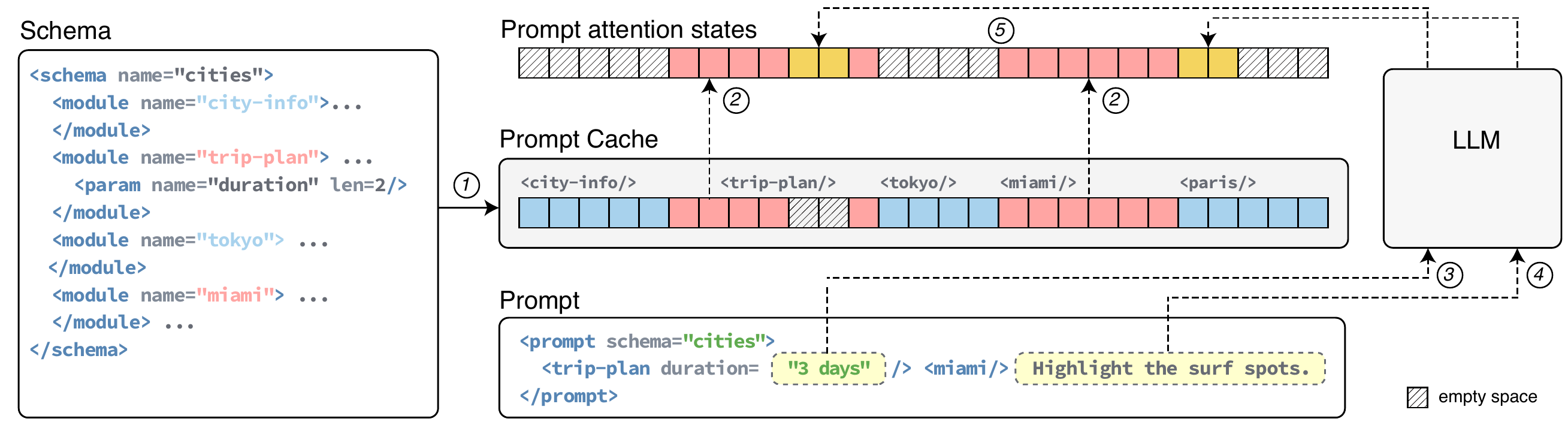}
\caption{Reuse mechanism in \tech: 
(\textit{i}) First, PML (\S\ref{ssec:schema}) makes reusable \modus explicit in both \Sch and Prompt.
A \modu can have parameters like \code{trip-plan}. A prompt importing the module supplies a value (\code{3 days}) to the parameter (\code{duration}). The prompt can include new text segments in place of excluded modules and parameters and at the end.
(\textit{ii}) Second, \modu encoding ($\S~\ref{ssec:encoding}$) precomputes attention states (\wcircled{1}) for all modules in the \sch and caches them for future reuse. 
(\textit{iii}) Third, when the prompt is served, \tech employs cached inference (\S\ref{ssec:cached-inference}): it retrieves the attention states cached for imported \modus (\wcircled{2}, computes them for parameters (\wcircled{3}) and new text segments (\wcircled{4}), and finally concatenates them to produce the attention states for the entire prompt (\wcircled{5}). 
This figure is an elaboration of Step \bcircled{1} in \autoref{fig:overview-sub3}.
}

\label{fig:schema-prompt}
\end{figure*}

\subsection{Overview}
The attention states of a text segment can only be reused if the segment appears at the same position in the LLM input.
This is because transformer architectures integrate unique positional embeddings into the $(k, v)$ attention states. 
This is not a problem for serving a single prompt using \kvcache, because the same prompt text is located at the same position, \ie, the beginning of the input, in all steps. 

Shared text segments, on the other hand, can appear in different positions in different prompts. To reuse their attention states across prompts, a caching system must tackle two problems. 
First, it must allow reuse despite a text segment appearing in different positions in different prompts.
Second, the system must be able to efficiently recognize a text segment whose attention states may have been cached in order to reuse, when the system receives a new prompt.

To tackle these two problems, we combine two ideas.
The first is to make the structure of a prompt explicit with a \emph{Prompt Markup Language} (\lang).
As illustrated by~\autoref{fig:schema-prompt}, the \lang makes reusable text segments explicit as modules, \ie, \emph{\modu}. 
It not only solves the second problem above but opens the door for solving the first, since each \modu can be assigned with unique position IDs. Our second idea is our empirical finding that LLMs can operate on attention states with discontinuous position IDs. As long as the relative position of tokens is preserved, output quality is not affected. This means that we can extract different segment of attention states and concatenate them to formulate new meanings. We leverage this to enable users to select \modus based on their needs, or even replace some meanings during the runtime.

\tech puts these two ideas together as follows. 
An LLM user writes their prompts in \lang, with the intention that they may reuse the attention states based on \modus. 
Importantly, they must derive a prompt from a \emph{\sch}, which is also written in \lang. \autoref{fig:schema-prompt} shows a example prompt based on an example \sch.
When \tech receives a prompt, it first processes its \sch and computes the attention states for its \modus. It reuses these states for the \modus in the prompt and other prompts derived from the same \sch.

We detail the design of \lang in \S\ref{ssec:schema} with a focus on techniques that maximize the opportunity of reusing. We explain how \tech computes the attention states of \modus in a \sch in \S\ref{ssec:encoding}, and how it may affect the output quality.
We explain how \tech reuse attention states from a \sch for the service of a prompt in \S\ref{ssec:cached-inference}.

The modular KV cache construction in \tech bears resemblance to the approximations observed in \emph{locally masked attention}~\cite{beltagy2023longformer,tay2023efficient}, which optimizes computations by setting a limited window for attention score calculations rather than spanning its attention across every token in its input sequence. Consider a scenario within \tech where each \modu is encoded independently. Given that attention states are strictly calculated within the confines of the \modu, this closely mirrors the setup of an attention mask that screens out sequences external to the \modu.
Therefore, the approximation made by \tech is to limit the attention window to each \modu. 
We note that employing such attention masks does not necessarily reduce output quality, as we will discuss in \S\ref{sec:eval}. In some contexts, these masks may even introduce beneficial inductive biases by effectively filtering out irrelevant information.

\subsection{Prompt Markup Language (\lang)}
\label{ssec:schema}

We next describe the key features of \lang that is used to define both \schs and \sch-derived prompts. 

\subsubsection{Schema vs. Prompt}
A \sch is a document that defines prompt modules and delineates their relative positions and hierarchies. Each \sch has a unique identifier (via the \code{name} attribute) and designates prompt modules with the \code{<module>} tag. Texts not enclosed by \code{<module>} tags or unspecified identifier are treated as anonymous prompt modules and are always included in prompts that are constructed from the schema.

For an LLM user, the \sch serves as an interface to create and reuse attention states for prompt modules. The user can construct a prompt from a \sch, with the \code{<prompt>} tag. 
This tag specifies the \sch to use through the \code{schema} attribute, lists the \modus to import, and adds any additional (non-cached) instructions.
 For example, to import the module \code{miami} from the \sch in \autoref{fig:schema-prompt}, one would express it as \code{<miami/>}.  
\tech will only compute the attention states for the text that is not specified in the \sch, e.g., \code{Highlights the surf spots} in \autoref{fig:schema-prompt}, and reuse attention states for the imported modules, e.g., \code{trip-plan} and \code{miami}, thereby reducing the latency.

\subsubsection{Maximizing Reuse with Parameters}
\label{ssec:parameter}
\lang allows a \modu to be parameterized in order to maximize the reuse opportunities. 
A parameter is a named placeholder with a specified length that can appear anywhere in a \modu in a \sch.
It is defined using the \code{<param>} tag, with the \code{name} and \code{len} attributes indicating its name and the maximum number of tokens for the argument, respectively.
When a prompt imports the \modu, it can supply a value to the parameter.
\autoref{fig:schema-prompt} shows an example of a paramterized \modu (\code{trip-plan}) and how a prompt would include the \modu and supply a value (\code{3 days}) to its argument (\code{duration}). Augment values are not cached.

There are two important uses of parameterized \modus.
First, it is common that a \modu differs from another only in some well-defined places. Parameters allow users to provide specific arguments to customize the module at runtime and still benefit from reusing. \autoref{fig:schema-prompt} illustrates this use case with \code{trip-plan}. This is especially useful for templated prompts.  Second, a parameter can be used to create a ``buffer" at the beginning or end of a \modu in the \sch. This buffer allows the user to add an arbitrary text segment in a prompt as long as the segment is no longer than the parameter token length it replaces.

\subsubsection{Other Features}
\textbf{Union modules}:~~Certain prompt modules exhibit mutually exclusive relationships. That is, within a set of modules, only one should be selected. For instance, consider a prompt that asks the LLM to suggest a book to read based on the reader's profile described by a \modu.
There could be multiple \modus each describing a reader profile but the prompt can include only one of them.
\begin{lstlisting}[style=xmlstyle]
<union>
  <module name="doc-en-US"> ... </module>
  <module name="doc-zh-CN"> ... </module>
</union>
\end{lstlisting}
To accommodate these exclusive relationships, we introduce the concept of a \textit{union} for prompt modules. A union of modules is denoted using the \code{<union>} tag. Prompt modules nested within the same union share the same starting position ID. A union not only streamlines the organization of the layout but also conserves position IDs used to encode prompt modules. Further, the system can utilize this structure for optimizations, such as prefetching.

While parameterized modules and unions appear to be similar, they are different in two aspects. First, as we will show in \S\ref{ssec:encoding}, parameters and union modules are encoded in different ways. Second, they serve different purposes: parameters are used for inline modifications to maximize the reuse of a module, while union modules are intended for better prompt structure and more efficient utilization of position IDs.

\textbf{Nested modules}:~~\lang also supports nested modules to express hierarchical prompt modules. That is, a \modu could include \modus or unions as components. In prompts, nested modules are imported as modules within modules as shown in \autoref{fig:parameterized}.

\textbf{Compatibility with LLM-specific template}:~~Instruction-tuned LLMs often adhere to specific templates to format conversations. For example, in Llama2, a single interaction between the user and the assistant follows the template: \code{<s>[INST] user message [/INST] assistant message </s>}. To reduce the effort required to manually format the prompt schema to match such templates for different LLMs, we introduce three dedicated tags: \code{<system>} for system-level prompts, \code{<user>} for user-generated prompts, and \code{<assistant>} for exemplar responses generated by the LLM. \tech dynamically translates and compiles these specialized tags to align with the designated prompt template of the LLM in use.

\subsubsection{Deriving \lang from Prompt Programs}
\label{ssec:deriving-schema}

To simplify \lang writing, \tech can automatically convert prompt programs~\cite{beurer2023prompting,Guidence} from languages like Python into \lang, eliminating the need for manual schema writing. This is primarily achieved using a Python API that transforms Python functions into corresponding \lang schemas. The conversion process is straightforward: if statements become \code{<module>} constructs in \lang, encapsulating the conditional prompts within. When a condition evaluates to true, the corresponding module is activated. Choose-one statements, such as if-else or switch statements, are mapped to \code{<union>} tags. Function calls are translated into nested prompt modules. Additionally, we have implemented a decorator to manage parameters, specifically to restrict the maximum argument length. This corresponds to the len attribute in the \code{<param>}. This Python-to-PML compilation hides \lang complexity from the user provides better maintainability of the prompt.

\subsection{Encoding \Sch}
\label{ssec:encoding}

The first time the attention states of a \modu are needed, they must be computed and stored in the device memory, which we refer to as \emph{prompt module encoding}. First, \tech extracts token sequences of a \modu from the schema. It then assigns position IDs to each token. The starting position ID is determined by the absolute location of the \modu within the schema. For instance, if two preceding prompt modules have token sequence sizes of 50 and 60 respectively, the \modu is assigned a starting position ID of 110. An exception exists for the union modules. Since prompt modules within the union start from the same positions, their token sequence size is considered with the size of the largest child.

From the token sequences of the \modu and the corresponding position IDs, these are then passed to the LLM to compute the $(k,v)$ attention states. We note that the assigned position IDs do not start from zero. This is semantically acceptable since white spaces do not alter the meaning of the precomputed text. However, many existing transformer positional encoding implementations, such as RoPE, often require adaptations to accommodate discontinuous position IDs, which we will discuss in ($\S~\ref{ssec:position_ids}$).

For encoding parameterized prompt modules, we use the idea that having white space in a prompt does not affect its semantics. Parameters are replaced by a predetermined number of \code{<unk>} tokens, equivalent to their \code{len} attribute value. The position IDs corresponding to these \code{<unk>} tokens are logged for future replacement.
When this module is integrated into a user's prompt and paired with the relevant arguments, the token sequences of these supplied arguments adopt the position IDs previously linked with the \code{<unk>} tokens. The resulting $(k,v)$ attention states then replace the attention states initially allocated for the \code{<unk>} tokens. We note that the length of the newly provided tokens can be smaller than the specified parameter length, as trailing white spaces do not change the semantics.

\textbf{Attention masking effect}: \tech confines attention score computation to the span of each prompt module, masking the attention states across modules. This masking effect can enhance or degrade output quality depending on the semantic independence of the modules. For semantically independent modules, masking reduces noise and improves quality. However, for semantically dependent modules, it can have the opposite effect. Therefore, each prompt module should be self-contained and semantically independent from other modules. One way to remove the masking effect is to use a method we refer to as \emph{scaffolding}. At the cost of additional memory, we allow users to specify ``scaffolds'', which are sets of prompt modules that are encoded together to share the attention span, in addition to their individual attention states. When all prompt modules in a scaffold are imported in a prompt, the attention states of the scaffold overrides the individual attention states. Scaffolding trades off additional memory for output consistency, which may be useful for applications that need deterministic results.

\subsection{Cached Inference}
\label{ssec:cached-inference}
When a prompt is provided to \tech, \tech parses it to ensure alignment with the claimed \sch. It verifies the validity of the imported modules. Then, as illustrated in Figure~\ref{fig:schema-prompt}, \tech retrieves the $(k, v)$ attention states for the imported \modus from the cache (\wcircled{2}), computes those for new text segments (\wcircled{3} and \wcircled{4}), and concatenates them to produce the attention states for the entire prompt (\wcircled{5}), replacing the prefill operation.

To detail the process, \tech starts by concatenating the KV state tensors corresponding to each imported \modu in the prompt. 
For instance, when a user prompt utilizes modules \(A,B\), the concatenated KV tensor is formulated as: $(k_C, v_C) = (\textrm{concat}(k_A, k_B), (\textrm{concat}(v_A, v_B))$.  It is worth noting that the order of concatenation does not matter due to the permutation invariance of transformers~\cite{dufter2022position}. This step solely requires memory copy.
Then, \tech computes the attention states for the segments of the prompt that are not cached, specifically, token sequences not defined in the \sch and arguments for parameterized \modus. \tech first identifies the position IDs of uncached texts based on their position relative to other utilized prompt modules. For example, if the text is situated between module A and B, it is assigned the position ID starting from the concluding positions of A, assuming gaps exist between the positions of A and B. Augments for parameterized \modus are assigned to the position IDs of \code{<unk>} tokens. Subsequently, the token sequences and position IDs are aggregated and passed to the LLM \emph{using $(k_C, v_C)$ as a \kvcache}, to compute the attention states for the entire prompt. It is important to note that the computational complexity for generating subsequent tokens remains consistent with that of \kvcache, as \modus are not employed beyond the initial token. In essence, \tech diminishes the latency involved in producing the first token, or time-to-first-token (TTFT).

\textbf{Memory optimization in batch inference}:
\label{ssec:batch}
Prompts are usually served in a batch for better GPU utilization. Different prompts derived from the same \sch may include the same prompt modules, such as system prompts. This opens up additional optimization opportunities by reducing \kvcache redundancies in a batch. Paged attention~\cite{kwon2023efficient} can resolve this issue by sharing the \emph{pointer} to the same prompt module across different prompts, instead of duplicating the attention states. Here, the use of \tech can implicitly improve system throughput by allowing more prompts to be processed in parallel.
\section{Implementation}
\label{sec:implementation}

We build a \tech prototype using the HuggingFace transformers library~\cite{wolf2020huggingfaces} in PyTorch and comprises 3K lines of Python code. We aim to seamlessly integrate with an existing LLM codebase and reuse its weights. We implement \tech to use both CPU and GPU memory to accommodate prompt modules and evaluate it on both platforms.

\subsection{Storing Prompt Modules in Memory}

We store encoded prompt modules in two types of memory: CPU memory (host DRAM) and GPU memory (HBM). To manage tensors across both memory types, we employ the PyTorch~\cite{pytorch} memory allocator.
Beyond simply pairing CPUs with prompt modules in CPU memory and GPUs with GPU memory, we also enable GPUs to access prompt modules stored in CPU memory. This is done by copying the prompt modules from the host to the device as needed. This process incurs a host-to-device memory copy overhead. Nonetheless, it allows the GPU to leverage the abundant CPU memory, which can scale up to terabyte levels. As we will show in \S\ref{sec:eval}, the computational savings from \tech more than compensate for the latencies caused by memory copy operations.
Using GPUs exposes trade-offs between memory capacity and latency: GPU memory is faster but limited in capacity, while CPU memory can scale easily yet incurs additional memory copy overhead. It appears feasible to contemplate a caching mechanism that leverages both CPU and GPU memory. We leave the development of a system that incorporates cache replacement and prefetching strategies to future research.

\subsection{Adapting Transformer Architectures}\label{ssec:position_ids}

Implementing \tech requires support for discontinuous position IDs (\S\ref{ssec:schema}). Although the Transformers library currently does not offer these features, they can be integrated with minor modifications. For instance, approximately 20 lines of additional code are needed for each LLM. We outline the required adjustments:

\textbf{Embedding tables:} Early models like BERT~\cite{vaswani2023attention} and GPT-2~\cite{radford2018improving} use lookup tables for mapping position IDs to learned embeddings or fixed bias, requiring no alterations.

\textbf{RoPE:} LLMs such as Llama2~\cite{touvron2023llama} and Falcon~\cite{falcon} adopt RoPE~\cite{jianlin2021roformer}, which employs rotation matrices for positional encoding in attention computations. We create a lookup table for each rotation matrix, enabling retrieval based on position IDs.

\textbf{ALiBi:} Utilized in models like MPT~\cite{MosaicML2023Introducing} and Bloom~\cite{scao2023bloom}, ALiBi~\cite{ofir2022train} integrates a static bias during softmax score calculations. Analogous to RoPE, we design a lookup table to adjust the bias matrix according to the provided position IDs.

We also override PyTorch's concatenation operator for more efficient memory allocation. PyTorch only supports contiguous tensors, and therefore, concatenation of two tensors always results in a new memory allocation. \tech needs to concatenate attention states of prompt modules, and the default behavior would lead to redundant memory allocations. We implement a buffered concatenation operator that reuses memory when concatenating tensors. This optimization improves the memory footprint of \tech and reduces the overhead of memory allocation.
\section{Evaluation}
\label{sec:eval}

Our evaluation of \tech focuses on answering the following three research questions:
(i) What is the impact of \tech on time-to-first-token (TTFT) latency and output quality (\S\ref{ssec:eval_latency} -- \S\ref{ssec:eval_latency_improve}), (ii) What is the memory storage overhead (\S\ref{ssec:eval_mem_overhead}), and (iii) What applications are a good fit for \tech (\S\ref{ssec:eval_usecase}). We use the regular \kvcache~\cite{pope2022efficiently} as our baseline. \tech and \kvcache share the exact same inference pipeline except for attention state computation. We use TTFT latency for comparison, which measures the time to generate the first token, as \tech and \kvcache have the same decoding latency after the first token.

\subsection{Evaluation Environment}\label{ssec:eval_env}

We evaluate \tech on two CPU configurations: an Intel i9-13900K accompanied by 128 GB DDR5 RAM at 5600 MT/s and an AMD Ryzen 9 7950X paired with 128 GB DDR4 RAM at 3600 MT/s. For our GPU benchmarks, we deploy three NVIDIA GPUs: the RTX 4090, which is paired with the Intel i9-13900K, and the A40 and A100, both virtual nodes hosted on NCSA Delta, each provisioned with a 16-core AMD EPIC 7763 and 224 GB RAM.
We employ several open-source LLMs, including Llama2, CodeLlama, MPT, and Falcon. We use LLMs that fit within the memory capacity of a single GPU (40 GB).
We utilize the LongBench suite~\cite{bai2023longbench} to assess TTFT improvements and output quality changes. LongBench encompasses a curated subsample of elongated data, ranging from 4K to 10K context length, excerpts from 21 datasets across 6 categories, including tasks like multi-document question answering~\cite{yang2018hotpotqa, ho2020constructing,trivedi2022musique,kovcisky2018narrativeqa,joshi2017triviaqa}, summarization~\cite{huang2021efficient,zhong2021qmsum,fabbri2019multi}, and code completion~\cite{guo2023longcoder,liu2023repobench}. We defined the documents in the LongBench datasets, such as wiki pages and news articles, as prompt modules. We kept the task-specific directives as uncached user text.

\subsection{Latency Improvements on Benchmark Datasets}\label{ssec:eval_latency}

\begin{figure}[t]
\includegraphics[width=0.47\textwidth]{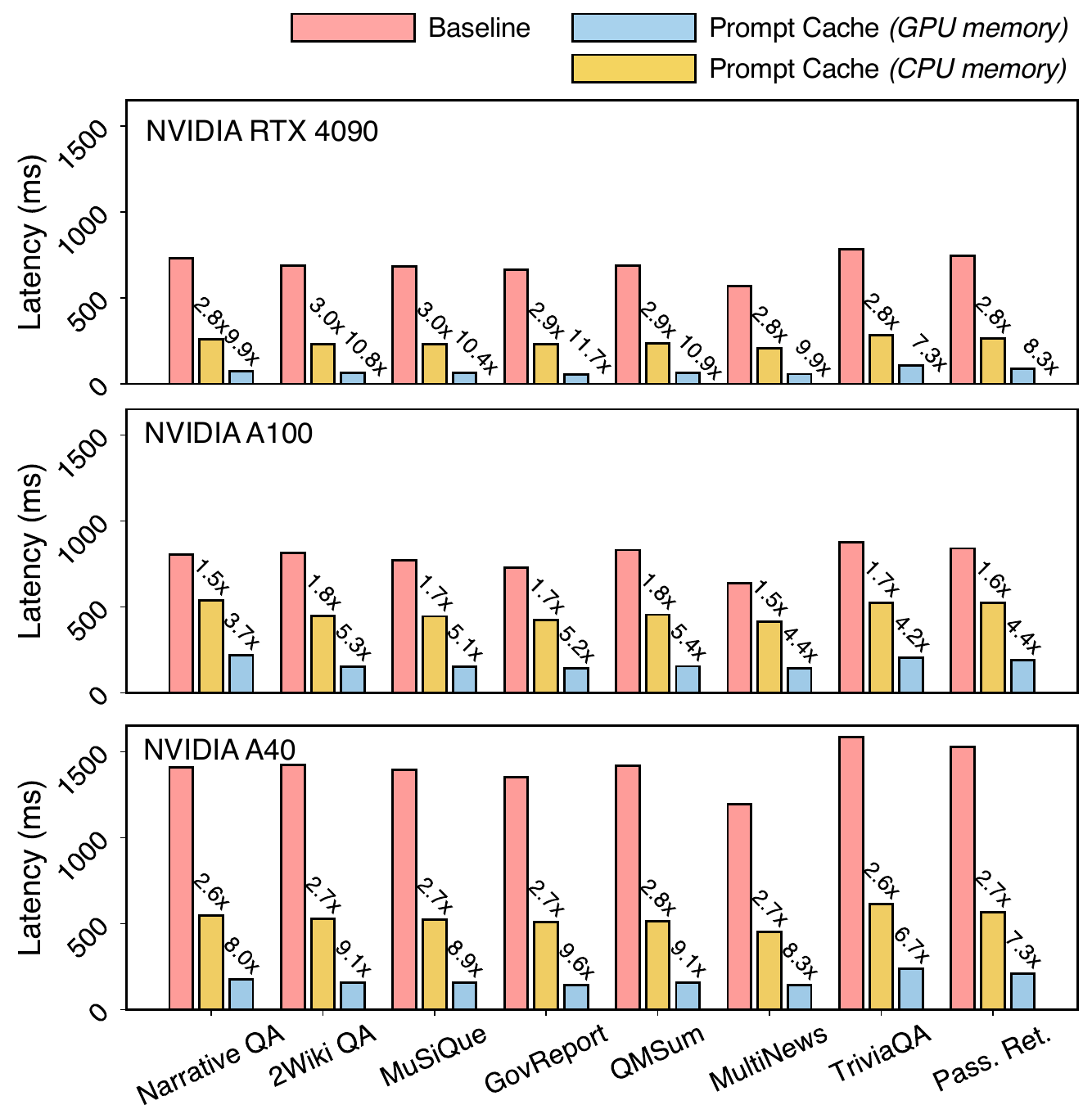}
\caption{GPU latency measurements: time-to-first-token (TTFT) for eight LongBench datasets across three NVIDIA GPUs.}
\label{fig:latency-gpu}
\end{figure}

\begin{figure}[t]
\includegraphics[width=0.47\textwidth]{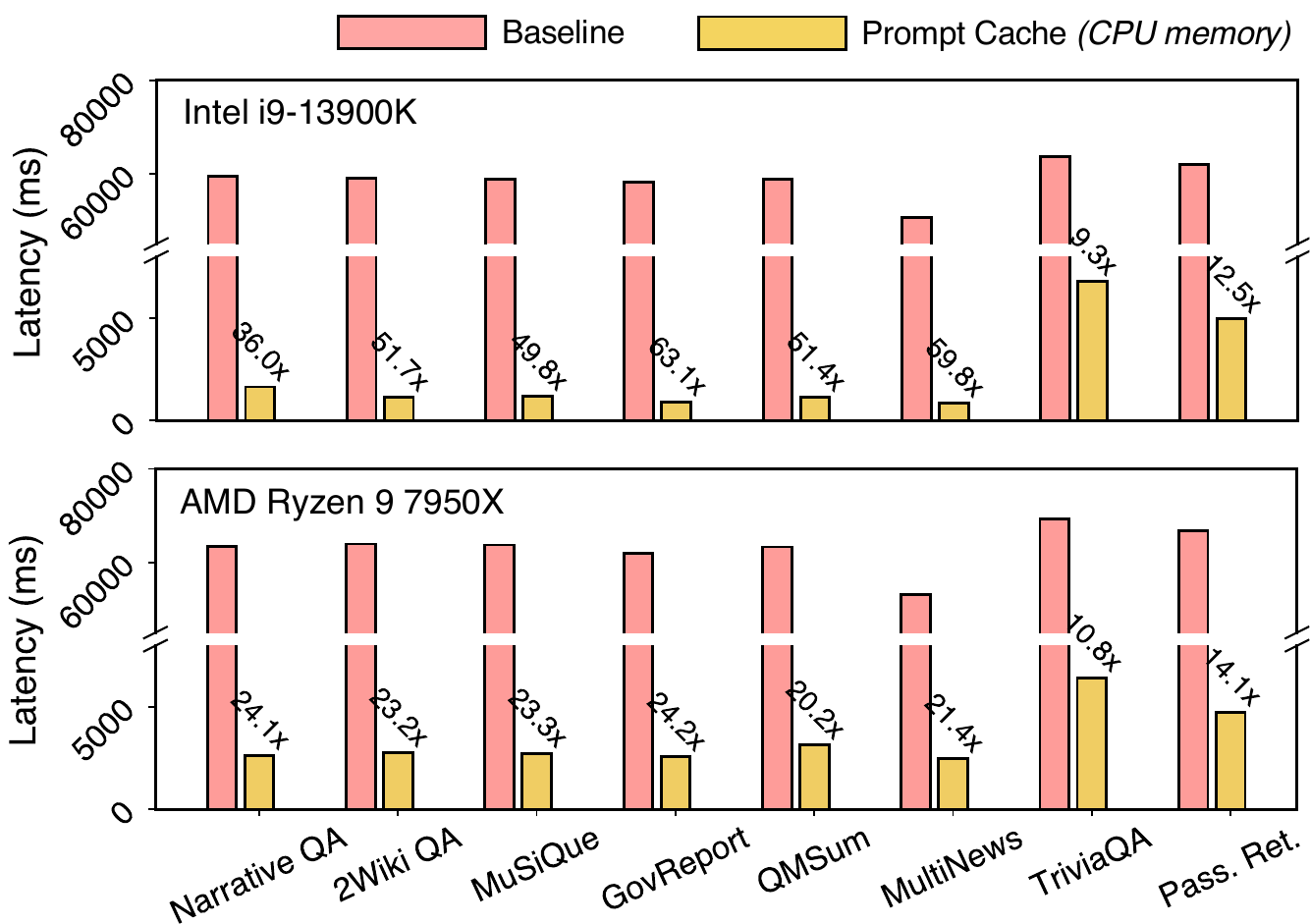}
\caption{CPU latency measurements: time-to-first-token (TTFT) for eight LongBench datasets across two CPUs.}
\label{fig:latency-cpu}
\end{figure}

We measured the TTFT latency on both GPU and CPU using Llama 7B, as shown in \autoref{fig:latency-gpu} and \autoref{fig:latency-cpu}. In our GPU evaluation, we used two memory setups: storing prompt modules in either CPU or GPU memory. For CPU experiments, we used CPU memory. Due to space constraints, we present only 8 benchmarks. The complete benchmark from 21 datasets can be found in the Appendix. 

\subsubsection{GPU Inference Latency}

We summarize our findings in \autoref{fig:latency-gpu}, evaluated on three NVIDIA GPUs: RTX 4090, A40, and A100. Yellow bars represent loading prompt modules from CPU memory, while blue bars represent the case in GPU memory. There is a consistent latency trend across the datasets since the LongBench samples have comparable lengths, averaging 5K tokens.
We observe significant TTFT latency reductions across all datasets and GPUs, ranging from 1.5$\times$ to 3$\times$ when using CPU memory, and from $5\times$ to 10$\times$ when employing GPU memory. These results delineate the upper and lower bounds of latency reductions possible with \tech. The actual latency reduction in practice will fall between these bounds, based on how much of each memory type is used.

\begin{table*}[ht!]
\centering
\small
\begin{tabular}{@{}c|c|ll|ll|ll|ll@{}}
\toprule
Dataset           & Metric  & \multicolumn{2}{c|}{Llama2 7B} & \multicolumn{2}{c|}{Llama2 13B} & \multicolumn{2}{c|}{MPT 7B} & \multicolumn{2}{c}{Falcon 7B} \\ \midrule
                  &         & Baseline       & Cached       & Baseline        & Cached       & Baseline      & Cached     & Baseline       & Cached       \\ \midrule
Narrative QA       & F1      & 19.93          & 19.38        & 20.37           & 19.94        & 10.43         & 11.33      & 7.14           & 8.87         \\
2 Wiki Multi-Hop QA   & F1      & \textbf{16.63}          & 13.95        & 14.59           & \textbf{17.69}        & 10.44         & \textbf{13.70}      & 14.42          & 15.07        \\
MuSiQue           & F1      & 7.31           & 8.57         & 10.03           & 12.14        & 7.38          & 7.32       & 4.81           & 5.86         \\
GovReport         & Rouge L & 24.67          & 25.37        & 28.13           & 28.18        & 26.96         & 27.49      & 22.39          & 23.40        \\
QMSum             & Rouge L & 19.24          & 19.46        & 18.80           & 18.82        & 15.19         & 15.51      & 12.84          & 12.96        \\
MultiNews         & Rouge L & 24.33          & 24.22        & 25.43           & 26.23        & 25.42         & 25.66      & 20.91          & 21.19        \\
TriviaQA          & F1      & 13.04          & 12.33        & 23.19           & 22.38        & 10.57         & 9.17       & 13.31          & 11.42        \\
Passage Retrieval & Acc     & \textbf{7.50}           & 4.25         & \textbf{9.08}            & 6.50         & 3.03          & 3.85       & 3.00           & 3.45         \\
\bottomrule
\end{tabular}
\caption{Accuracy benchmarks on LongBench datasets. We mark the outliers as \textbf{bold}, of which the performance is higher than $2.5$ compared to the counterpart.}
\label{table:accuracy}
\end{table*}

\subsubsection{CPU Inference Latency}

\autoref{fig:latency-cpu} shows that \tech achieves up to a $70\times$ and $20\times$ latency reduction on the Intel and AMD CPUs, respectively. We surmise that this disparity is influenced by the difference in memory bandwidth in system setups (5600MT/s DDR5 RAM on the Intel CPU versus 3600MT/s DDR4 RAM on the AMD CPU). As expected, the latency is higher for the datasets with a larger proportion of uncached prompts, such as \texttt{TriviaQA}.
Interestingly, CPU inference benefits more significantly from \tech than GPU inference does. This is attributed to the much greater latency of attention computation in the CPU, especially as the sequences become longer (\eg, lower FP16/FP32 FLOPs compared to GPU). This indicates that \tech is particularly beneficial for optimizing inference in resource-constrained environments, such as edge devices or cloud servers with limited GPU resources.

\subsection{Accuracy with \tech} \label{ssec:eval_accuracy}
To verify the impact of \tech on the quality of LLM response, without scaffolding, we measure accuracy scores with the LongBench suite. To demonstrate general applicability, we apply \tech to the three LLMs having different transformer architectures (\S\ref{ssec:position_ids}): Llama2, MPT, and Falcon.
The accuracy benchmark results shown in \autoref{table:accuracy} demonstrate \tech preserves the precision of the output. We use deterministic sampling where the token with the highest probability is chosen at every step so that the results with and without \tech are comparable. Across all datasets, the accuracy of output with \tech is comparable to the baseline. 

\begin{figure}[t]
    \includegraphics[width=0.485\textwidth]{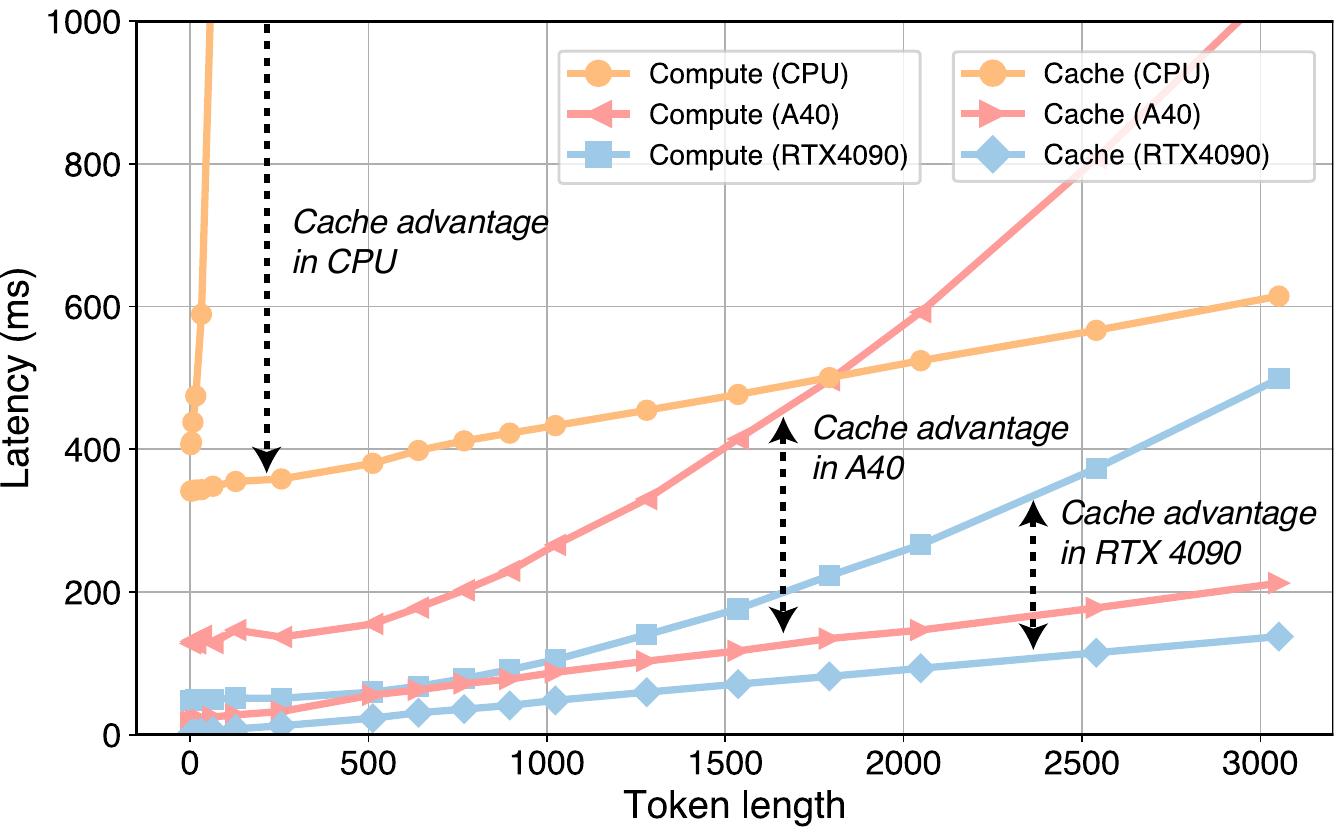}
    \caption{Cache advantage: A comparison of computational and caching overheads in GPUs and CPUs. While attention computation cost increases quadratically, the attention state memory copy overhead (\ie, \tech) rises linearly. Here, GPUs load prompt modules directly from CPU memory.}
    \label{fig:cache-advantage}
\end{figure}

\begin{table}
    \centering
    \small
    \begin{adjustbox}{max width=0.48\textwidth}
    \begin{tabular}{@{}lllll@{}}
        \toprule
        LLM      & BERT    & Falcon 1B  & Llama 7B  & Llama 13B   \\ \midrule
        MB/token & 0.03    & 0.18       & 0.50      & 0.78        \\ \midrule
        LLM      & MPT 30B & Falcon 40B & Llama 70B & Falcon 180B \\ \midrule
        MB/token & 1.31    & 1.87       & 2.5       & 4.53  \\ \bottomrule    
    \end{tabular}
    \end{adjustbox}
    \caption{Memory overhead of caching a single token}
    \label{table:memory-overhead}
    \end{table}

\subsection{Understanding Latency Improvements}\label{ssec:eval_latency_improve}

Theoretically, \tech should offer quadratic TTFT latency reduction over regular \kvcache. This is because, while \tech's memcpy overhead grows linearly with sequence length, computing self-attention has quadratic computational complexity with respect to sequence length. To validate this, we tested \tech on a synthetic dataset with varied sequence lengths, assuming all prompts were cached. We compared the TTFT latency of \tech to that of regular \kvcache using an Intel i9-13900K CPU and two GPUs (NVIDIA RTX 4090 and A40) with the Llama2 7B model. For both CPU and GPU, CPU memory is used for prompt module storage.

\textbf{Quadratic improvement}:
Our findings, presented in \autoref{fig:cache-advantage}, show that \kvcache's latency increases quadratically with sequence length, while \tech's memory copy cost grows linearly. This means that the latency advantage of \tech (the gap between the two curves) expands quadratically with sequence length. This difference is more pronounced on CPUs than GPUs since CPUs experience higher attention computation latencies, whereas the disparity between \tech's overhead, \ie, host-to-device memcpy in GPUs and host-to-host memcpy in CPUs is not significant. With attention states with 5K tokens, latency for host-to-host, host-to-device, and device-to-device memcpy are respectively 3.79 ms, 5.34 ms, and 0.23 ms.

\textbf{Effect of model size}: Furthermore, as the model's parameter size grows, so does the computational overhead for \kvcache. For example, moving from a 7B to 13B model at a token length of 3K added 220 ms latency, whereas \tech added only 30 ms. This difference stems from the fact that LLM complexity also scales quadratically with hidden dimension size. For example, the FLOPS of attention is $6nd^2+4n^2d$, for prefill operation. This suggests that \tech's advantage over \kvcache also quadratically increases with model size (\ie, hidden dimension).

\textbf{End-to-end latency}: Since \tech reduces only TTFT, its impact on the time needed to receive the complete LLM response diminishes as the number of generated tokens increases. 
For instance, on the RTX 4090 with Llama 7B for 3K context, \tech enhances TTFT from 900 ms to 90 ms, while the token generation time or the time-to-subsequent-token (TTST) remains consistent between \kvcache and \tech at an average of 32 ms per token, regardless of the token length.
Nonetheless, a quicker response time contributes positively to the user experience and the overall end-to-end latency~\cite{lew2018interactivity, liu2023cachegen}, For instance, Given that \tech enhances TTFT from 900 ms to 90 ms, this equates to the generation of 25 more tokens within the same timeframe.
Another factor is that \tech enables sharing attention states within the same batch, as we discussed in \S\ref{ssec:batch}. Depending on the workload characteristics, \tech can improve overall throughput by utilizing the larger batch size enabled by the reduced memory footprint. For example, suppose there are 100 requests, each with a 2K token prompt. If all prompts share the same 1K token module, \tech can reduce the memory footprint by 50\% when combined with methods like paged attention, allowing for a larger working batch size and thus higher throughput.

\subsection{Memory Overhead}\label{ssec:eval_mem_overhead}

The memory overhead associated with \tech is proportional to the aggregated number of tokens cached. This overhead can be determined by referencing both the prompt schema and the target LLM. In \autoref{table:memory-overhead}, we elucidate the memory overhead on a per-token basis, under the assumption of utilizing a 16-bit precision for floating points.
For compact models, such as Falcon 1B, caching a document containing 1K tokens would require approximately 180 MB of memory. If there are hundreds of prompt modules, the combined memory consumption would range in the tens of gigabytes---a quantity within the memory confines of server-grade GPUs. Conversely, for larger models like Llama 70B, caching a 1K length module would command a substantial 2.5 GB of memory per document, which leaves CPU memory as the only option for prompt module storage. Given these considerations, compression methods for attention states~\cite{zhang2023h2o} remain an avenue for future research in prompt caching techniques.

\subsection{Applications of \tech}\label{ssec:eval_usecase}

\begin{figure}
    \includegraphics[width=0.49\textwidth]{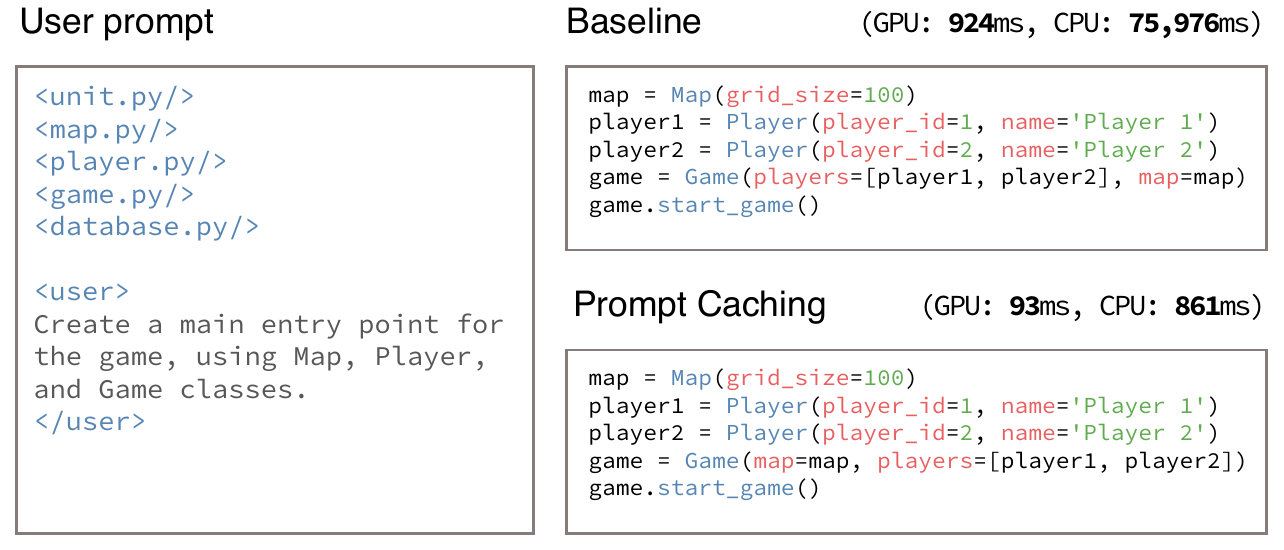}
    \caption{Code generation using \tech: Each source file becomes a prompt module, allowing users to ``import'' files in their prompt context with minimal overhead.}
    \label{fig:code-gen}
\end{figure}

\begin{figure}
    \includegraphics[width=0.49\textwidth]{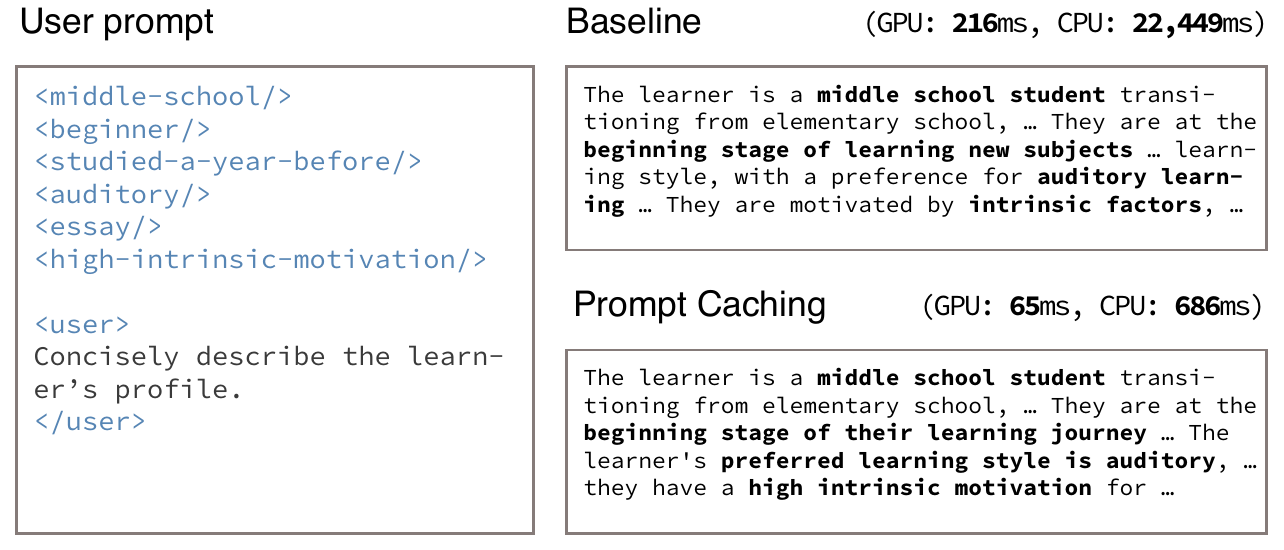}
    \caption{Personalization example: Six categories each have five traits. Traits in the same category are grouped in \texttt{<union>}.}
    \label{fig:personalization}
\end{figure}

\begin{figure}
    \includegraphics[width=0.49\textwidth]{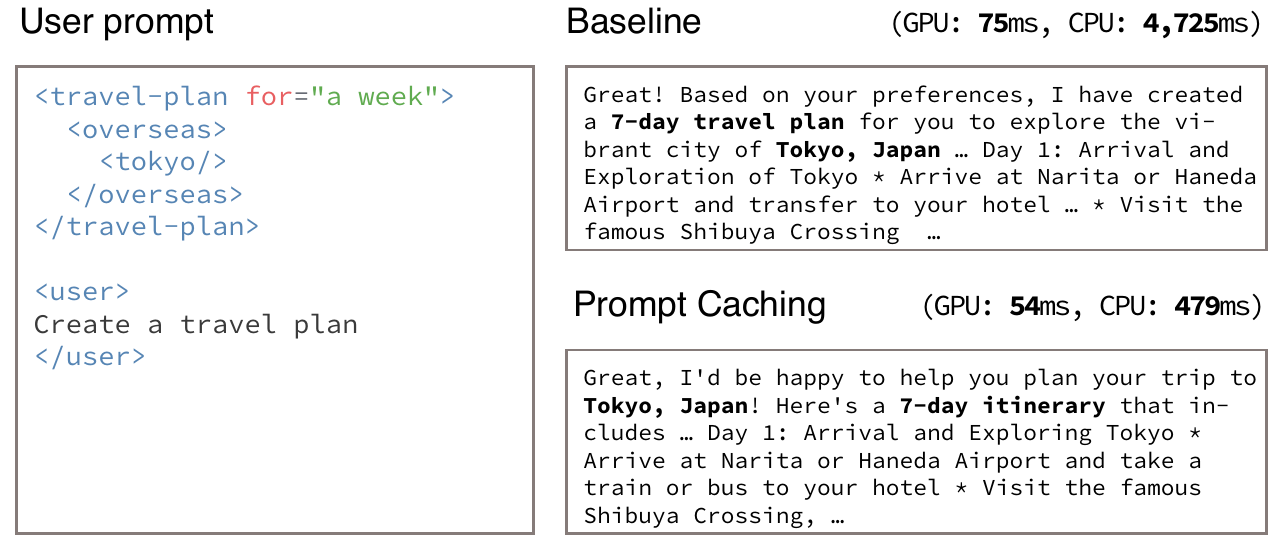}
    \caption{Parameterized prompts: The \texttt{<travel-plan>} is reconfigured at runtime while maintaining caching efficiency, offering flexible prompt structuring.}
    \label{fig:parameterized}
\end{figure}

We demonstrate the expressiveness of \lang with example use cases that require more complicated prompt structures and advanced features (\S\ref{ssec:schema}) than the LongBench benchmarks: (\textit{i}) multiple modules in a query, (\textit{ii}) union, and (\textit{iii}) parameterization. Furthermore, these tasks underscore the notable latency reduction as the number of cached tokens increases in such complicated use cases. Across use cases, we provide a qualitative assessment of the output by juxtaposing cached and non-cached generation, showcasing that \tech maintains output quality, along with the latency reductions achieved by \tech. We use Llama2 7B and store prompt modules in the local memory (\ie, GPU memory for GPU inference). The full schema for these tasks is available in Appendix B. 

\subsubsection{Code Generation}

LLMs are commonly used for code generation~\cite{guo2023longcoder, liu2023repobench}, aiding programmers in either assisting with or directly generating code. Currently available methods, such as Copilot~\cite{copilot}, typically focus on individual source files. \tech, however, can extend this to multiple files leveraging a modular nature of source code. For instance, each class or function could be a distinct prompt module.
\autoref{fig:code-gen} illustrates multi-source code generation using CodeLlama 7B~\cite{codellama}. We treat individual classes like \code{Unit}, \code{Map}, \code{Game}, and \code{Player} as prompt modules in our schema for game programming. Users can then include these prompt modules whenever they need them in the code. There is a $4\times$ improvement in TTFT latency on GPUs while the output remains identical.

\subsubsection{Personalization}

\autoref{fig:personalization} shows the latency benefits and the output quality of \tech in a personalization use case. Personalization is integral to many recommender systems~\cite{wu2023survey}, finding prominent applications in LLM contexts such as education, content recommendations, and targeted marketing.
We highlight the efficacy of feature-based personalization through \tech. Here, personalization hinges on a defined set of features. Each feature is represented as a distinct prompt module, with relationships between features denoted using union tags such as grade level, proficiency, learning history, learning style, and assessment type.

\subsubsection{Parameterized Prompts}

In \autoref{fig:parameterized}, we show a trip planning use case leveraging parameterization (\S\ref{ssec:schema}). The schema used in this use case encompasses one adjustable parameter to specify the trip duration along with two union modules to select the destination. Users can reuse the templated prompt with custom parameters, enjoying lower TTFT latency and the same quality of LLM response enabled by \tech.
\section{Conclusions and Future Work}

We introduce \tech, an acceleration technique based on the insight that attention states can be reused across LLM prompts. \tech utilizes a prompt schema to delineate such reused text segments, formulating them into a modular and positionally coherent structure termed ``prompt modules''. This allows LLM users to incorporate these modules seamlessly into their prompts, thereby leveraging them for context with negligible latency implications. Our evaluations on benchmark data sets indicate TTFT latency reductions of up to 8$\times$ on GPUs and 60$\times$ on CPUs.

For future work, we plan on using \tech as a building block for future LLM serving systems. Such a system could be equipped with GPU cache replacement strategies optimized to achieve the latency lower bound made possible by \tech. Different strategies for reducing host-to-device memory overhead can also be beneficial, such as the integration of compression techniques in the KV cache, or utilization of grouped query attention. Another promising exploration GPU primitives for sharing attention states across concurrent requests, as we breifly discussed in \S\ref{ssec:batch}. This can not only reduce the TTFT latency but also time-per-output-token (TPOT) latency by packing more requests into a single batch.
Finally, \tech can directly accelerate in-context retrieval augmented generation (RAG) methods, where the information retrieval system basically serves as a database of prompt modules. \tech can be particularly useful for latency-sensitive RAG applications in real-time question answering and dialogue systems.
\section*{Acknowledgements}
This work is supported in part by NSF Athena AI Institute (Award \#2112562), NSF Award \#2047220, and Yale University.  This work used the Delta system at the National Center for Supercomputing Applications (NCSA) through allocation CIS230289 from the Advanced Cyberinfrastructure Coordination Ecosystem: Services \& Support (ACCESS) program, which is supported by National Science Foundation grants (Award \#2138259, \#2138286, \#2138307, \#2137603, and \#2138296).
\newpage
\bibliography{main}
\bibliographystyle{mlsys2024}

\appendix


\end{document}